\documentclass[10pt,twocolumn]{article}

\usepackage{iccv}
\usepackage[table]{xcolor}
\usepackage{times}
\usepackage{epsfig}
\usepackage{graphicx}
\usepackage{amsmath}
\usepackage{amssymb}
\usepackage{bm} 
\usepackage{multirow,bigdelim}
\usepackage{pbox}
\usepackage{textcomp}
\usepackage{cite}
\usepackage{tabularx}
\usepackage{comment}

\usepackage{pgfplots}
\pgfplotsset{compat=1.14}
\usetikzlibrary{shapes.geometric, arrows}

\newcommand\approxtilde{\raise.17ex\hbox{$\scriptstyle\sim$}}

\definecolor{orange}{RGB}{242, 150, 73}
\definecolor{green}{RGB}{127,255,0}
\definecolor{blue}{RGB}{127,255,255}
\definecolor{purple}{RGB}{230, 135, 255}
\definecolor{yellow}{RGB}{251, 237, 65}
\definecolor{red}{RGB}{255, 40, 40}
\pgfplotscreateplotcyclelist{cycle-bar}{
	{black, thick, fill=blue}, 
    {black, thick,fill=purple}, 
    {black, thick, fill=yellow},
    {black, thick, fill=red}
}
\pgfplotscreateplotcyclelist{cycle-graph}{
	{blue!80!black, thick, mark=x, mark size=3, mark options={solid, blue!50!black}}, 
    {purple!80!black, thick, mark=triangle*, mark options={solid, purple!50!black}}, 
    {yellow!80!black, thick, mark=*, mark options={yellow!50!black}},
    {red!80!black, thick, mark=square*, mark options={red!50!black}},
    {green!80!black, thick, mark=diamond*, mark options={green!50!black}},
    {orange!80!black, thick, mark=pentagon*, mark options={orange!50!black}}
}

\usepackage[pagebackref=true,breaklinks=true,letterpaper=true,colorlinks,bookmarks=false]{hyperref}

\iccvfinalcopy 

\def\iccvPaperID{142} 

\ificcvfinal\fi
\begin{document}

\title{DeNet: Scalable Real-time Object Detection with Directed Sparse Sampling}

\author{Lachlan Tychsen-Smith, Lars Petersson\\
CSIRO (Data61)\\
7 London Circuit, Canberra, ACT, 2601\\
{\tt\small Lachlan.Tychsen-Smith@data61.csiro.au, \tt\small Lars.Petersson@data61.csiro.au}
}

\maketitle
\begin{abstract}
We define the object detection from imagery problem as estimating a very large but extremely sparse bounding box dependent probability distribution. Subsequently we identify a sparse distribution estimation scheme, Directed Sparse Sampling, and employ it in a single end-to-end CNN based detection model. This methodology extends and formalizes previous state-of-the-art detection models with an additional emphasis on high evaluation rates and reduced manual engineering. We introduce two novelties, a corner based region-of-interest estimator and a deconvolution based CNN model. The resulting model is scene adaptive, does not require manually defined reference bounding boxes and produces highly competitive results on MSCOCO, Pascal VOC 2007 and Pascal VOC 2012 with real-time evaluation rates. Further analysis suggests our model performs particularly well when finegrained object localization is desirable. We argue that this advantage stems from the significantly larger set of available regions-of-interest relative to other methods.  Source-code is available from: \href{url}{https://github.com/lachlants/denet}
\end{abstract}

\section{Introduction}

Feed-forward neural networks exhibit good convergence properties given a random initialization under stochastic gradient descent (SGD) and, given an appropriate network design and training regime, can generalize well to previously unseen data\cite{lenet}. In particular, convolutional neural networks (CNNs) built from interleaved convolution and pooling layers with ReLU activation functions have set numerous benchmarks in computer vision tasks\cite{lenet}\cite{resnet}\cite{googlenet}. A number of methodologies have been developed to map their state-of-the-art dense regression and classification capabilities to the problem of identifying axis aligned bounding boxes of object instances in images. In particular we highlight the relatively slow \textit{region} based CNN approaches (R-CNN\cite{rcnn}, Faster R-CNN\cite{faster-rcnn}) and the more recent work on real-time detection (YOLO\cite{yolo}, SSD\cite{ssd}). 

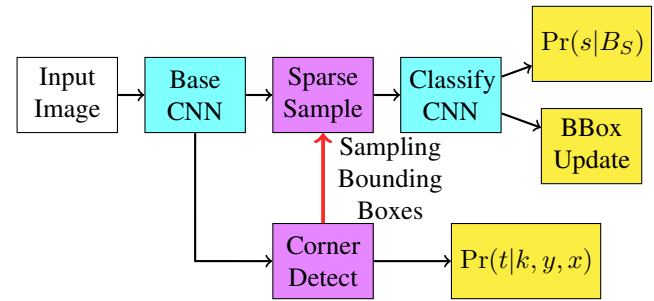
\begin{figure}[tb]
\centering
\begin{tikzpicture}[scale=1.7]
\tikzstyle{block} = [rectangle, text centered, draw=black, fill=white, minimum size=1cm, text width=1.1cm,  anchor=center]
\node (image) [block] at (0,0) {Input Image};
\node (base) [block, fill=blue] at (1,0) {Base CNN};
\node (dss) [block, fill=purple] at (2,0) {Sparse Sample};
\node (cls) [block, fill=blue] at (3,0) {Classify CNN};
\node (bbox) [block,  fill=yellow] at (4.1,-0.4) {BBox Update};
\node (pr_cls) [block, text width=1.35cm, fill=yellow, xshift=0.175cm] at (4,0.4) {$\mathrm{Pr}(s|B_S)$};
\node (corner) [block,  fill=purple] at (2,-1.3) {Corner Detect};
\node (pr_corner) [block,text width=1.75cm,anchor=west, fill=yellow] at (3,-1.3) {$\mathrm{Pr}(t|k,y,x)$};
\draw [->,thick] (image) -- (base);
\draw [->,thick] (base) -- (dss);
\draw [->,thick] (dss) -- (cls);
\draw [->,thick] (cls) -- (pr_cls);
\draw [->,thick] (cls) -- (bbox);
\draw [->,thick] (base) -- (1,-1.3) -- (corner);
\draw [->,thick] (corner) -- (pr_corner);
\draw [->,line width=0.5mm,red] (corner) -- node[right, text centered, text width=1.5cm, black,pos=.5] {Sampling Bounding Boxes} ++ (dss);
\end{tikzpicture}
\caption{A high level flow diagram depicting the DeNet methodology. The CNN's are highlighted in blue, the novel components in purple and the outputs in yellow. The sampling bounding box dependency $B_S$ (highlighted in red) is held constant during back propagation to produce an end-to-end trained model. The corner distribution and final classification distribution are jointly optimized using cross entropy loss. }
\end{figure}
\newpage 
Rather than focusing on obtaining state-of-the-art accuracy in a competition environment (i.e. computationally unconstrained) in this paper we emphasis the dual task of obtaining the best detection performance at a predefined evaluation rate i.e. 60 Hz and 30 Hz. The primary contributions made within this paper include:
\begin{itemize}
\item An improved theoretical understanding of modern detection methods and a generic framework in which to describe them i.e. Directed Sparse Sampling.
\item A novel, fast, region-of-interest estimator which doesn't require manually defined reference bounding boxes.  
\item A novel application of deconvolution layers which greatly improves evaluation rates. 
\item Six implementations of our method demonstrating competitive detection performance on a range of benchmarks. 
\item An easily extended Theano based code release to facilitate the research community.
\end{itemize}

\subsection{Related work} \label{sec:related_work}

In region based CNN detection (R-CNN)\cite{rcnn} the image is first preprocessed with a region proposal algorithm e.g. selective search\cite{ssearch}, region proposal network (RPN) \cite{faster-rcnn}, etc. This algorithm identifies image regions (i.e. bounding boxes) of \textit{interest} (RoIs) which are then rescaled to fixed dimensions (normalizing scale and aspect ratio) and fed into a CNN based classifier. The CNN assigns a probability that the region bounds an object of interest or the \textit{null} class and, via linear regression, identifies an improved bounding box. This approach has demonstrated state-of-the-art results, however, it is very expensive to train and evaluate, requiring multiple full CNN evaluations (one per region proposal) and an often expensive pre-processing step. Since the majority of CNN computation occurs in the first few layers, Fast R-CNN\cite{fast-rcnn} addressed these issues by applying a shallow CNN to the image and then, for each region, extracting fixed sized features from the generated feature map for the final classification. In Faster R-CNN\cite{faster-rcnn} the region proposal algorithm was integrated into the CNN providing an end-to-end solution, improved timings and demonstrating that both tasks (region proposal and classification) shared similar underlying features. Despite these improvements, to our knowledge, region based CNN's have not been demonstrated operating near real-time frequencies.

In You Only Look Once (YOLO)\cite{yolo} they depart from the algorithmically defined \textit{region} based approaches described above, opting instead for a predefined, regular grid of detectors. In effect they merged the region classification problem into the region proposal network (RPN) first proposed in Faster R-CNN. With this approach the CNN is only evaluated once to produce the outcomes for all detectors resulting in significantly reduced training and evaluation times. In Single Shot Detector (SSD)\cite{ssd} this approach was further refined with an improved network design and training methodology to demonstrate comparable results to the region based methods. We note that the considerable improvements achieved with SSD required scene dependent engineering to manually predefine the most likely set of regions within the image to contain an object, a flaw shared with the Faster R-CNN region proposal network. In particular, SSD demonstrated an improvement of $2.7\%$ MAP\cite{ssd} by the addition of four aspect ratios to the predefined regions on the Pascal VOC2007\cite{pascal-voc} dataset, highlighting the importance of manual engineering in modern state-of-the-art detector designs. Without going into too much detail, we note that in practice manually engineered solutions typically limit scalability and adaptiveness to different problem sets (without an expensive re-engineering process).

The primary differentiator between these methods lies in how each method identifies and treats the \textit{regions} to be classified. R-CNN based methods sample regions sparsely based on an algorithmic preprocessing step and normalize the region of interest while YOLO based approaches perform dense sampling with a manually defined grid of detectors without image normalization. Often dense methods are well suited to current implementations and, therefore offer a significant timing advantage over sparse methods. However, in this work, we demonstrate a novel model design which combines the ease of training, scene adaptability and classification accuracy of the sparse region-based approaches with the fast training and evaluation of the dense non-region based methods. 

\subsection{Probabilistic Object Detection}

We formulate the probabilistic multiclass detection problem as first estimating the distribution $\mathrm{Pr}(s | B, I)$ where $s \in C \cap \lbrace \mathrm{null} \rbrace$ is a 
random variable indicating the presence of an instance of class $c \in C$ or the \textit{null} class (indicating no instances) which is \textit{sufficiently} bounded by the box $B = \lbrace x, y, w, h \rbrace$ and $I$ is the input image (omitted in subsequent derivations).
This formulation incorporates the assumption that only a single instance of a class can occupy each bounding box.
We note that this definition does not seek to perform instance assignment, but can be used as an input to an algorithm that does e.g. Non-Max Suppression.

Given a suitable neural network design we assert that $\mathrm{Pr}(s | B)$ can be estimated from training data with class bounding box annotations. However, since the number of unique bounding boxes is given by $|B| \propto XYWH$ where $(X,Y)$ are the number of image positions and $(W,H)$ the range of bounding box dimensions the naive solution quickly becomes intractable. For instance, assuming the most common settings for the ImageNet dataset, $1000$ classes and $224\times 224$ images, and considering all valid bounding boxes within the image, expressing this distribution requires approximately $629 \times 10^{9}$ values or 2.5TB in 32bit float format. Clearly this is an intractable problem with current hardware. 

At the cost of localization accuracy, subsampling the output bounding boxes is a valid approach. 
For instance, by careful dataset dependent manual engineering, Faster R-CNN and YOLO based approaches subsample the distribution to the order of $10^4$ to $10^5$ bounding boxs \cite{yolo} \cite{faster-rcnn}. These boxes are then refined by estimating only the most likely bounding box in a local region via linear regression. As an alternative to large scale subsampling, we sought to exploit the fact that, due to occlusion and other factors, we expect a very small subset of bounding boxes to contain class instances other than the \textit{null} class. Subsequently, we have developed a solution based on the state-of-the-art regression capabilities of a single end-to-end CNN which estimates the highly sparse distribution $\mathrm{Pr}(s | B)$ in a real-time (or computationally constrained) operational environment.

\section{Directed Sparse Sampling (DSS)} \label{sec:dss}

We use the term \textit{Directed Sparse Sampling} to refer to the method of a applying a jointly optimized two stage CNN where one stage estimates the likely locations where user-defined \textit{interesting} values occur and the other sparsely classifies the identified values e.g. in R-CNN based models (including R-FCN and DeNet) we estimate the bounding boxes which are most likely to include a non-\textit{null} class assignment, then run a classifer over these bounding boxs. 

\subsection{Corner-based RoI Detector}
Here we introduce the concept of bounding box corner estimation for efficient region-of-interest (RoI) estimation. In our methodology, this task is performed by estimating the likelihood that each position in the image contains an instance of one of 4 corner types i.e. $\mathrm{Pr}(t | k,y,x)$ where $t$ is a binary variable indicating the presence of a corner of type $k \in \lbrace \mathrm{top~left}, \mathrm{top~right}, \mathrm{bottom~left}, \mathrm{bottom~right} \rbrace$ at position $(x,y)$ in the input image. We assert that due to the natural translation invariance of the problem, estimating the corner distribution can be efficiently performed with a standard CNN design trained on bounding box annotated image data (e.g. MSCOCO\cite{mscoco}, Pascal VOC\cite{pascal-voc}, etc). 

With the corner distribution defined we estimate the likelihood that a bounding box $B$ contains an instance by applying a Naive Bayesian Classifier to each corner of the bounding box:

\begin{gather}
\label{eq:bbox_bayes}
\begin{split}
\mathrm{Pr}(s \neq \mathrm{null} | B) \propto \prod_k \mathrm{Pr}(t|k,y_k,x_k) 
\end{split}
\end{gather}

where $(x_k,y_k) = f_k(B)$ indicates the bounding box position associated with each corner type $k$. For ease of implementation we define the $N \times N$ bounding boxes 
with the largest non-null probability $\mathrm{Pr}(s \neq \mathrm{null} | B)$ as the \textit{sampling bounding boxes} $B_S$. The user defined variable $N$ balances the maximum number of detections the model can handle with the computational and memory requirements. 

With the potentially non-null bounding boxes estimated, we pass a feature vector of predefined length from the corner detector model to the final classification stages. Therefore, the final classification stage is a function of the form $f:\bar{\alpha}_{B} \rightarrow \mathrm{Pr}(s | B)$ where $\bar{\alpha}_{B}$ is a feature vector uniquely identified by the sampling bounding box $B \in B_S$.  It is important that the feature is uniquely associated with each bounding box, otherwise the classifier will have no information to distinguish between bounding boxes with the same $\bar{\alpha}_B$. Exactly how to construct the feature vector is still a matter of debate\cite{r-fcn,faster-rcnn} however we construct $\bar{\alpha}_{B}$ by concatenating together the \textit{nearest neighbour} \textit{sampling} features at predefined locations relative to each sampling bounding box (e.g. bounding box corners, center, etc) in addition to the bounding box width and height. The bounding box center position was omitted  from the feature vector such that the classifier would be agnostic to image offsets.

\subsection{Training}

During training, the model is initially forward propagated to generate the sampling bounding boxes $B_S$ as described in the previous subsection. In addition, we augment the sampling bounding boxes with the ground truth bounding boxes and randomly generated samples. We then propagate the activations $\bar{\alpha}_B$ associated with the augmented set of sampling bounding boxes through the rest of the model to produce the final classification distribution $\mathrm{Pr}(s | B_S)$ and updated bounding box parameters. The set of sampling bounding boxes $B_S$ is held constant during gradient estimation to enable end-to-end training, therefore the corner detector network is optimized in conjunction with the bounding box classification and estimation task. Since forward propagation is a necessary preprocessing step in the back propagation based SGD policy typically used to optimize neural networks, the DeNet method introduces no penalty to training time over a standard dense network. 

The DeNet model jointly optimizes over the corner probability distribution, final classification distribution and bounding box regression cost, i.e.

\begin{gather}
\label{eq:cost}
\begin{split}
\mathrm{Cost} =& \frac{\lambda_t}{\Lambda_t} \sum_{k,y,x} \phi(t | k,y,x) \ln(\mathrm{Pr}(t | k,y,x) ) + \\ & \frac{\lambda_s}{\Lambda_s} \sum_{B \in B_S} \phi(s| B) \ln(\mathrm{Pr}(s|B) ) +  \\ &
\frac{\lambda_b}{\Lambda_b} \sum_i \mathrm{SoftL_1}(\phi_{B,i} - \beta_i)
\end{split}
\end{gather}

where $\phi(...)$ are the ground truth corner and classification distributions, $\phi_{B,i} = \lbrace x_i,y_i,w_i,h_i \rbrace$ the ground truth bounding boxes, $(\lambda_s,\lambda_t,\lambda_b)$ are user defined constants indicating the relative strength of each component, $(\Lambda_s,\Lambda_t,\Lambda_b)$ are constants normalizing each component to 1 given the model initialization and $\mathrm{SoftL_1}(x)$ is defined in \cite{fast-rcnn}. The corner distribution $\phi(t | k,y,x)$ is identified by mapping each groundtruth instance's corners to a single position in the corner map, corners out of bounds are simply discarded. The detection distribution $\phi(s| B)$ is identified by calculating the \textit{intersection over union} (IoU) overlap between the groundtruth bounding boxes and the sampling bounding boxes $B_S$. Following standard practice, the regression target bounding box $\phi_B$ is identified by selecting the ground truth bounding box with the largest IoU overlap. 

\newpage

\subsection{Detection Model} \label{sec:model}

Residual neural networks \cite{resnet} have demonstrated impressive regression capabilities on a number of large scale datasets. In particular the 101 layer Residual Network model (ResNet-101) achieved state of the art performance on the ILSVRC2015\cite{imagenet} and MSCOCO\cite{mscoco} datasets when combined with Faster R-CNN. As the base model to our networks we selected the 34 layer, 21M parameter ResNet-34 model (DeNet-34) and the 101 layer, 45M parameter ResNet-101 model (DeNet-101).


To each base model we modified the input size to $512\times 512$ pixels, removed the final mean pooling and fully connected layers and appended two deconvolution\cite{deconv} layers followed by a corner detector. The corner detector is responsible for generating the corner distribution and produces a \textit{feature sampling map} via a learnt linear projection with $F_s$ features at each spatial position. The deconvolution\cite{deconv} layers efficiently reintroduce spatial information that was lost in the base model such that the feature map and corner probability distribution can be defined at a greater spatial resolution i.e. $64\times 64$ compared to $16\times 16$ without. This results in a $16\times 16$ pixel minimum size for each sampling bounding box.

Following the corner detector is the \textit{sparse layer} which observes the corners identified by the corner detector and generates a set of sampling bounding boxes (RoIs). The RoIs are used to extract a set of $N \times N$ feature vectors from the \textit{feature sampling maps}. In this case, we are sparsely sampling $N^2$ bounding boxes from a set of 4.2M valid bounding boxes. A feature vector is constructed by extracting the \textit{nearest neighbour} sampling features associated with a $7\times 7$ grid plus the bounding box width and height. This produces a feature with $7 \times 7 \times F_s + 2$ values. We found that nearest neighbour sampling was sufficient because the feature sampling maps have the same, relatively high, spatial resolution as the bounding box corners. Finally, the feature vectors are propagated through a relatively shallow fully connected network to generate the final classification and fine tuned bounding box for each sampling RoI. 

\begin{table}[tb] 
\begin{center}
\resizebox{0.475\textwidth}{!}{
\begin{tabular}{ l|r|r|r|r|r|r|r|r}
Model & $F_0$ & $F_1$ & $F_2$ & $F_3$ & $F_4$ & $F_5$ & $F_6$ & $F_7$\\
\hline
DeNet-34 & 512 & 256 & 128 & 4706 &1536 &1024&768 &512\\
DeNet-101 & 2048 & 384 & 192 & 6274 & 2048 & 1536 &1024&768\\
\end{tabular}}
\end{center}
\caption{Filter parameters used for DeNet models. See Table \ref{table:model_desc}}
\label{table:model_filters}

\begin{center}
\begin{tabular}{ l | r | r | r | r}
Layer & Input Shape &  Filters & Shape & Stride\\
\hline
\multicolumn{5}{c}{
ResNet-34 or ResNet-101\cite{resnet} base model.} \\
\hline
Deconv & $16\times 16\times F_0$ & $F_1$ & $3\times 3$ & $2\times 2$ \\
Deconv & $32\times 32\times F_1$ & $F_2$ & $3\times 3$ & $2\times 2$ \\
Corner & $64\times 64\times F_2$ & - & -  &- \\
Sparse & - & - &- & - \\ 
Conv  & $N\times N\times F_3$ & $F_4$ & $1\times 1$  & $1\times 1$ \\ 
Conv  & $N\times N\times F_4$ & $F_5$ & $1\times 1$  & $1\times 1$ \\ 
Conv & $N\times N\times F_5$ & $F_6$ & $1\times 1$  & $1\times 1$ \\ 
Conv  & $N\times N\times F_6$ & $F_7$ & $1\times 1$  & $1\times 1$ \\ 
Classifier & $N\times N\times F_7$ & - & - & - \\
\end{tabular}
\end{center}
\caption{DeNet: A ResNet derived model for DSS Object Detection with a $512 \times 512$ input image. Layers in the base models above the line are initialized with a pretrained ResNet-34 or ResNet-101 ImageNet 2012 classification model.}
\label{table:model_desc}
\end{table}

In Table \ref{table:model_filters} and \ref{table:model_desc} we describe the additional layers appended to the base models with the following definitions:
\begin{itemize}
\item \textbf{Conv}: Convolves a series of 2D filters over the input activations. Filter weights were initialized via the normal distribution $\mathcal{N}(0,\sigma)$ with $\sigma^2 = 2 / (n_f n_x n_y)$ where $n_f$ is the number of filters and $(n_x, n_y)$ their spatial shape \cite{init}. Following each convolution is batch normalization \cite{batchnorm} then the ReLU activation function.
\item \textbf{Deconv}: Applies a learnt deconvolution\cite{deconv} (upsampling) operation followed by ReLU activation. In this case it is equivalent to upscaling both spatial dimensions then applying a \textbf{Conv} layer. 
\item \textbf{Corner}: Estimates a corner distribution via the softmax function and produces a sampling feature map. See Section \ref{sec:dss}.
\item \textbf{Sparse}: Identifies sampling bounding boxes from corner distribution and produces a fixed size sampling feature from the sampling feature maps. 
\item \textbf{Classifier}: Maps activations to the desired probability distribution via the softmax function and generates bounding box targets.
\end{itemize}

For DeNet-34 we use a ResNet-34 base model and $F_s=96$ to produce a feature vector of $4706$ values and a total of 32M parameters. The DeNet-101 model uses a ResNet-101 base model and increased the number of filters by approximately $1.5\times$ for the appended layers (See Table \ref{table:model_filters}). These changes produce a sparse feature vector of $6274$ values and a total of 69M parameters.

\subsubsection{Skip Layer Variant}

As an extension we considered augmenting the DeNet models with \textit{skip} layers. In recent work, \textit{skip} layers have demonstrated consistent improvements in classification\cite{resnet}, detection\cite{fpn} and semantic segmentation\cite{semanticskip} and, more generally, are an integral component to highway \cite{highway} and residual networks\cite{resnet}. In this case, these layers connect the \textbf{Deconv} layer with the final layer in the base model which has the same spatial dimensions. Our implementation follows \cite{fpn}, each skip layer performs a linear projection of the source features to the destination feature dimensions and simply adds the resulting feature maps (before activation). 
 
\subsubsection{Wide Variant}

In this model we modified the skip model variants to use a $128\times 128$ spatial resolution for the corner and feature sampling maps by the addition of another \textbf{Deconv} and skip layer. We also increased $N$ to $48$ to produce 2304 RoIs. In the current implementation, this approach comes with a considerable timing cost due to the increased classification burden and the CPU bound algorithm for identifying RoIs. With further engineering (e.g. deduplication) we believe these costs could be reduced.

\section{Implementation Details} \label{sec:implement} 

Our models are implemented within our Theano based CNN library called DeNet. The source-code is available from: \href{url}{https://github.com/lachlants/denet}

\subsection{Training methodology} \label{sec:training}

In all experiments we used Nesterov style SGD\cite{nesterov-sgd} with an initial learning rate of 0.1, 
momentum of 0.9 and weight decay of 0.0001 (only applied to weights). A batch size of 128 was employed for both models with 32 samples per GPU iteration. The learning rate was divided by 10 at epoch 30 and epoch 60 and a total of 90 training epochs were performed. Note that, apart from the batch size changes, these hyperparameters are identical to those used when training the original residual networks for classification \cite{resnet}. No online hard negative mining \cite{ohem} or other gradient optimization techniques were applied, however, we observed some instances of overtraining on Pascal VOC. In response, to increase exposure to negative samples, we introduced $10\%$ randomly generated bounding box samples during training.

An augmentation strategy very similar to GoogLeNet\cite{googlenet} was employed  to improve model generalization to different scales and translations. For each sample, a black border was added to the smallest dimension to produce a square image. At test time, this image was scaled to $512 \times 512$ pixels using bilinear sampling, during training a random crop was selected with an area between $(0.08,1.0)$ relative to the border image and an aspect ratio between $( 3/4, 4/3 )$. The random crop was discarded and a new one generated if no ground truth objects overlapped with the crop by at least 50\%. This process was repeated up to 10 times and, as a fallback, the entire bordered image was returned. As in testing, the resulting crop was scaled to $512 \times 512$ pixels. Random photometric (contrast, saturation and brightness) and mirror augmentation was also employed \cite{googlenet}. 

\subsection{Identifying Sampling Bounding Boxes (RoIs)} \label{sec:implement_sample_bbox} 

A simple algorithm was developed to quickly search the corner distribution for non-null bounding boxes:
\begin{enumerate}
\item Search the corner distribution for corners $\lbrace k,y,x \rbrace \in C_{\lambda}$ where $\mathrm{Pr}(t=1 | k,y,x) > \lambda$. 
\item For each corner type, select the $M$ corners with the greatest likelihood $C_M \subseteq C_{\lambda}$
\item Generate a set of unique bounding boxes by matching every corner within $C_M$ of type $\mathrm{top-left}$ with every one of type $\mathrm{bottom-right}$.
\item Calculate the probability of each bounding box being non-null via Equation \ref{eq:bbox_bayes}.
\item Repeat steps 2 and 3 with corners of type $\mathrm{top-right}$ and $\mathrm{bottom-left}$.
\item Sort bounding boxes by probability and keep the $N^2$ largest to produce the sampling bounding boxes $B_S$.
\end{enumerate}
Since the vast majority of corners are culled in step 1 this method obtains a significant speed up beyond the naive brute force method i.e. testing every possible bounding box. 

\section{Results and Analysis}

In this section we compare our design with previously published models. We note that in some cases, an \textit{apples-to-apples} comparison is difficult due to the wide range of base models, data augmentation schemes and dataset merging. In particular, we note that SSD utilize larger batch sizes while R-CNN models have larger input resolutions (on average). All our DeNet timing results are provided for a single Titan X GPU (CuDNN 5110) with a batch size of 8x, the same settings used in SSD. For brevity we include only three flavours of the non real-time Faster R-CNN model, the original RPN (VGG), the ResNet-101 extension RPN+ (ResNet-101) and R-FCN for comparison (highlighted in grey in the tables). We note that due to implementation restrictions RPN based models are tested with a single image per batch.

\begin{table}[htb]
\begin{center}
\begin{tabular}{ l |c| c| c|c }
Model & Max. Input & BS & L & Param. \\
\hline
\rowcolor[gray]{.85} RPN (VGG)\cite{faster-rcnn} & $1000\times 600$ & 1 & 16 & 137M \\
\rowcolor[gray]{.85} RPN+ (ResNet)\cite{resnet} & $1000\times 600$ & 1 & 100 & 45M \\
\rowcolor[gray]{.85} R-FCN \cite{r-fcn} & $1000\times 600$ & 1 & 100 & 45M \\
Fast YOLO\cite{yolo} & $448\times 448$ & 1 & 9 & 9M  \\
YOLO\cite{yolo} & $448\times 448$ & 1 & 26 & 60M \\
SSD300\cite{ssd} & $300\times 300$ & 8 & 25 & 27M  \\
SSD512\cite{ssd} & $512\times 512$ & 8 & 25 & 27M \\
\hline
DeNet-34 (ours) & $512\times 512$ & 8 & 41 & 33M \\
DeNet-101 (ours) & $512\times 512$ & 8 & 107  & 69M \\
\end{tabular}
\end{center}
\caption{Model overview detailing maximum input image sizes, batch size at test time (BS), number of activation layers (L) and approximate number of parameters. }
\label{table:model_overview}
\end{table}

In Table \ref{table:model_overview} we provide a broad overview of the baseline models. We note that despite an increased number of layers and parameters the DeNet models obtain improved evaluation rates (See Section \ref{sec:timing_breakdown}). 

\subsection{Hyperparameter Optimization}

For the following we used the DeNet-34 model and trained it on Pascal VOC 2007 \texttt{train} and Pascal VOC 2012 \texttt{trainval} (14,041 images), for testing we used Pascal VOC 2007 \texttt{val} (2,510 images). The same DeNet-34 model initialization was used for all experiments. We applied the training procedure described in Section \ref{sec:training} except with a batch size of 96. 

\begin{table}[tb] 
\setlength{\extrarowheight}{1pt}
\begin{center}
\resizebox{0.475\textwidth}{!}{
\begin{tabular}{ c | c | c | c | c | c |c|c }
$\bm{\lambda_s}$ & 1 & 1 & 1 & 1 & 1 & 1 & 1 \\
$\bm{\lambda_t}$ & 1 & 10 & 10 & 100 & 100 & 100 & 500\\
$\bm{\lambda_b}$ & 1 & 1 & 10 & 0.1 & 1 & 10 & 1\\
\hline
\textbf{MAP (\%)} & 49.6 & 63.3 & 64.9 & 71.8 & 72.8 & 72.1 & 70.6 \\
\end{tabular}}
\end{center}
\caption{Optimizing cost hyperparameters $\lbrace \lambda_s, \lambda_t, \lambda_b \rbrace$, see Equation \ref{eq:cost}. MAP is provided for Pascal VOC 2007 \texttt{val} dataset.}
\label{table:optimize_lambda}
\end{table}

\begin{table}[t] 
\setlength{\extrarowheight}{1pt}
\begin{center} 
\resizebox{0.475\textwidth}{!}{
\begin{tabular}{l|r|r|r|r|r|r|r}
\textbf{Sample BBoxs} & 64 & 144 & 256 & 400 & 576 & 784 & 1024\\ 
\textbf{Coverage (\%)} & 81 & 87 & 91 & 93 & 95 & 96 & 97\\
\textbf{Eval. Rate (Hz)} & 96 & 90 & 84 & 76 & 69 & 61 & 55 \\
\end{tabular}}
\end{center} 
\caption{Sample bounding boxes vs coverage over training dataset and evaluation rate.}
\label{table:cover_n}
\end{table}

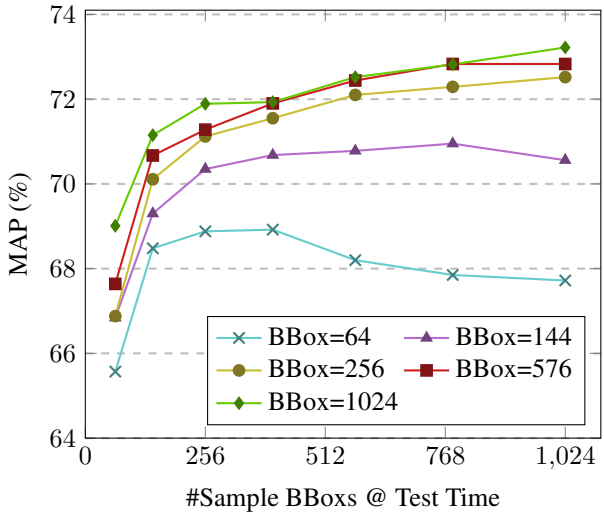
\begin{figure}[tb] 
\begin{tikzpicture}
	\begin{axis}[
        grid style={thick, dashed},
		xlabel=\#Sample BBoxs @ Test Time, xtick={1024,768,512,256,0},
		xmin=0, xmax=1100,
		ymajorgrids=true,
		ylabel=MAP (\%), ymin=64, ymax=74.1,
        cycle list name=cycle-graph,
        legend pos=south east,
		legend cell align=left,
        legend columns=2
	]
	\addplot coordinates {(64,65.57) (144,68.48) (256,68.88) (400,68.92) (576,68.20) (784,67.85) (1024,67.72)};
	\addplot coordinates {(64,66.84) (144,69.30) (256,70.35) (400,70.68) (576,70.78) (784,70.95) (1024,70.56)};
	\addplot coordinates {(64,66.88) (144,70.11) (256,71.12) (400,71.55) (576,72.10) (784,72.29) (1024,72.52)};
	\addplot coordinates {(64,67.64) (144,70.67) (256,71.28) (400,71.90) (576,72.44) (784,72.83) (1024,72.83)};
	\addplot coordinates {(64,69.01) (144,71.15) (256,71.89) (400,71.93) (576,72.52) (784,72.82) (1024,73.22)};
	\legend{BBox=64, BBox=144, BBox=256, BBox=576, BBox=1024}
	\end{axis}
\end{tikzpicture}
\caption{The MAP on the Pascal VOC 2007 validation dataset with varying number of bounding box samples during training (see legend) and testing (displayed on x-axis).}
\label{fig:scale_n}
\end{figure}
In Table \ref{table:optimize_lambda} we performed a coarse search over the corner and bounding box regression cost parameters $\lambda_t$ and $\lambda_b$. Best results were achieved by setting $\lambda_s=1$, $\lambda_t=100$ and $\lambda_b=1$, these were applied in all subsequent experiments. Next, we investigated the model behaviour with varying numbers of sampling bounding boxes. In particular, we trained a set of models with $N=\lbrace 8, 12, 16, 24, 32 \rbrace$.  At test time, we took each of these models and varied $N$ from 8 to 32 to produce Figure \ref{fig:scale_n}. In Table \ref{table:cover_n}, we provide the model evaluation rate and \textit{coverage} (percentage of ground truth with a sampling bounding box with $\mathrm{IoU} > 0.5$) over the training set that was obtained by the RoI estimator described in Section \ref{sec:dss}. As expected, we observed a consistently improving MAP with diminishing returns above 576 when training with a larger number of sampling bounding boxes. In general we observed an improved MAP with increased testing bounding boxes at the cost of evaluation rates. For subsequent experiments we set $N=24$ for both training and testing.

\subsection{Timing Breakdown and Evaluation Rates} \label{sec:timing_breakdown}

In Table \ref{table:timing_breakdown} we present a coarse analysis of the timing for both DeNet models. We broke the timing into 4 sequentially executed stages:
\begin{enumerate}
\item \textbf{Estimate corners}: Images are uploaded to the GPU and fed through the base network generating the corner distribution and sampling feature maps. The corner distribution is transferred from GPU to CPU memory. 
\item \textbf{Generate RoI}: The sampling bounding boxes (RoIs) are generated from the corner distribution. 
\item \textbf{Classify RoI}: The final classification CNN is executed, the classification distribution and bounding box regression outputs are transferred from GPU to CPU. 
\item \textbf{Estimate instances}: Non-Max Supression is run over the resulting detection \textit{hits} producing a de-duplicated list of detections for each image. 
\end{enumerate}

\begin{table}[tb]
\begin{center}
\begin{tabular}{rrrrr}
 & \multicolumn{2}{l}{\textbf{DeNet-34}} & \multicolumn{2}{l}{\textbf{DeNet-101}}\\
Estimate corners& 8.4 ms & 68\% & 24.9ms & 80\%  \\
Generate RoI & 1.5 ms & 12\% & 1.5 ms & 5\% \\
Classify RoI & 2.4 ms & 19\% & 4.5 ms & 15\% \\
Estimate instances & 0.1 ms & 1\% & 0.1 ms & 0\% \\
\hline
Total (per image) & 12.4 ms & - & 31.0 ms & - \\
\end{tabular}
\end{center}
\caption{A coarse timing breakdown per image for DeNet models at test time on a Titan X GPU.}
\label{table:timing_breakdown}
\end{table}

We observe that the vast majority of time is spent evaluating the base network to generate corners. Also, note that the CPU bound \textbf{Generate RoI} stage timing can vary substantially between different samples and may require additional tuning depending on application. Furthermore, we wish to emphasize a number of important features of the DeNet model which makes it significantly faster than most other baseline models: 

\begin{itemize}
\item \textbf{Deconvolution}: Spatial information is increased via deconvolution layers as opposed to the \textit{atrous} modified models used in R-FCN and SSD. This method introduces spatial information significantly later in the model, greatly improving evaluation rates. 
\item \textbf{Fast RoI Features}: Features are extracted via a simple \textit{nearest neighbour} sampling method, limiting the the number of feature reads to 49 per RoI. Some RPN variants use pooling which varies from 49-580 per RoI. 
\item \textbf{Input Image Dimensions}: DeNet scales all images to 512x512 pixels, whereas RPN based methods use a varying input size up to 1000x600 pixels.  
\item \textbf{Batching}: Our models are tested with 8x samples per batch (same as SSD). This improves GPU utilization. 
\end{itemize}

With these improvements in timing we are able to use a more expressive base model for the same evaluation rate. 

\subsection{RoI Coverage Comparison}
 
In Table \ref{table:roi_quality}. we provide the coverage obtained by by the top 300 RoIs for RPN, R-FCN and DeNet methods. We observe that given a relatively low number of RoIs, RPN (VGG) and R-FCN provide better coverage at low IoU thresholds, however with increasing IoU the DeNet models provide significantly improved coverage.

\begin{table}[htb]
\begin{center}
\begin{tabular}{l|c|c|c|c|c}
 & \multicolumn{5}{c}{\textbf{Top 300 Coverage@IoU} (\%)}\\
Model & 0.5 & 0.6 & 0.7 & 0.8 & 0.9\\
\hline
\rowcolor[gray]{.85} RPN (VGG) & 89.24 & 81.99& 65.18 & 27.79 & 2.66\\
\rowcolor[gray]{.85} R-FCN & 91.41 & 86.72 & 77.37 & 45.37 & 5.73\\
DeNet-34 & 80.98 & 77.00 & 71.63 & 63.39 & 46.37 \\
DeNet-101 & 82.47 & 78.69 & 73.81 & 65.36 & 47.83\\
\end{tabular}
\end{center}
\caption{Coverage on Pascal VOC 2007 \texttt{test} using 300 sample bounding boxs (RoI proposals).}
\label{table:roi_quality}
\end{table}

We note that RPN / R-FCN utilize bounding box regression and deduplication methods in their RoI proposal networks, these factors improve coverage with low numbers of proposals. As demonstrated in the following sections, the DeNet RoI coverage results do not necessarily translate to a reduced MAP for the full model, which includes NMS and bounding box regression, at lower IoU thresholds.  

\subsection{MSCOCO} The Microsoft Common Object in Context\cite{mscoco} dataset consists of 82K training and 40K validation images distributed across 80 classes. For testing, the dataset includes an 80K test dataset from which a user known subset of 20K images forms the \texttt{test-dev2015} set and an unknown subset of 20K images forms \texttt{test2015} allowing only 5 evaluations. Due to the dataset size, number of classes and relatively small size of the object instances within the images, MSCOCO is a considerably more difficult dataset compared to the Pascal VOC challenges. The primary evaluation metric for MSCOCO is the integral of the MAP over the detection matching parameter IoU=0.5 to IoU=0.95. This metric places a greater emphasis on localization performance compared to the Pascal datasets. We found that setting $\lambda_t = 50$ for DeNet-101 was necessary for convergence, this is likely due to the greater number of corners present within each image on average compared to the validation experiments. Training took \approxtilde 4 days with $2\times$ Tesla P100 GPUs for DeNet-34 and \approxtilde 6.5 days with $4\times$ Tesla P100 GPUs for DeNet-101.

In Table \ref{table:mscoco} we provide the precision and recall results for our models on \texttt{test-dev2015}. The DeNet models demonstrate a clear advantage over other high evaluation rate implementations e.g. our real-time DeNet-34 model beats SSD300 by $6.2\%$ MAP at the same evaluation rate and SSD512 by $2.6\%$ at more than twice the evaluation rate. The DeNet-101 model furthers this advantage and is only beaten by the very slow \textit{competition} style RPN+ model utilizing multi-scale evaluation and bounding box refinement. At the time of writing, the DeNet-101 model obtains a result good enough to be in the top-10 on the MSCOCO competition leaderboard which doesn't consider evaluation time.
The \textit{skip} model variants consistently improved performance on small and medium sized objects (see AR and AP for small and medium area objects in table)
with a minor cost to large objects and evaluation rate. The \textit{wide} variants further improved small object detection and fine object localization at the cost of evaluation rate. Near identical results were obtained on MSCOCO \texttt{test-std2015} e.g. we obtained a MAP@[0.5:0.95] of 29.3\% and 31.7\% for DeNet-34 and DeNet-101 respectively. Analysis suggests our advantage stems from improved large object detection and \textit{finegrain} object localization performance, as reflected in the MAP@IoU=0.75 result. We argue this is an outcome of the much larger range of candidate RoI's our method produces e.g. the vanilla DeNet models can select from a possible set of $4.2\times 10^6$ bounding boxes while SSD utilizes $2.5\times 10^4$. Utilizing such a large set of candidate bounding boxes would likely be intractable with the dense evaluation methods used in the YOLO and RPN derived models. 


\begin{table*}[tb]
\begin{center}
\resizebox{\textwidth}{!}{
\begin{tabular}{l|c|ccc|ccc|ccc|ccc}
 & Eval. & \multicolumn{3}{|c|}{AP@IoU (\%)} & \multicolumn{3}{|c|}{AP@Area (\%)} & \multicolumn{3}{|c|}{AR@Dets  (\%)} & \multicolumn{3}{|c}{AR@Area (\%)} \\
 Model & Rate &0.5:0.95 & 0.5 & 0.75 & S & M & L & 1 & 10 & 100 & S & M & L\\
\hline
\rowcolor[gray]{.85} RPN (VGG) & 7 Hz & 21.9 & 42.7 & - & - & - & - & - & - & - & - & - & -\\
\rowcolor[gray]{.85} RPN+ (ResNet) & \textless 1  Hz & \textbf{34.9} & \textbf{55.7} & - & \textbf{15.6} & \textbf{38.7} & \textbf{50.9} & - & - &- &- & -&-\\
\rowcolor[gray]{.85} R-FCN & 9  Hz & 29.9 & 51.9 & - & 10.8 & 32.8 & 45.0 & - & - &- &- & -&-\\
SSD300 & 58  Hz & 23.2 & 41.2 & 23.4 & 5.3 & 23.2 & 39.6 & 22.5 & 33.2 & 35.3 & 9.6 & 37.6 & 56.5 \\ 
SSD512 & 23 Hz & 26.8 & 46.5 & 27.8 & 9.0 & 28.9 & 41.9 & 24.8 & 37.5 & 39.8 & 14.0 & 43.5 & 59.0 \\ 
\hline
DeNet-34 & \textbf{83 Hz}  & 29.4 & 46.2 & 31.2 & 7.8 & 30.8 & 47.4 & 26.9 & 38.0 & 38.5 & 11.2 & 41.9 & 63.0 \\
DeNet-34 (skip) & 82 Hz & 29.5 & 47.9 & 31.1 & 8.8 & 30.9 & 47.0 & 26.9 & 38.0 & 38.6 & 13.2 & 41.7 & 61.6 \\
DeNet-34 (wide) & 44 Hz & 30.0 & 48.9 & 31.8 & 10.1 & 30.9 & 45.7 & 27.3 & 39.5 & 40.3 & 17.0 & 42.8 & 60.9 \\
DeNet-101 & 34 Hz &  31.9 & 50.5 & 34.2 & 9.7 & 34.9 & 50.6 & 28.4 & 39.8 & 40.3 & 13.1 & 44.8 & 64.1 \\
DeNet-101 (skip) & 33 Hz & 32.3 & 51.4 & 34.6 & 10.5 & 35.1 & \textbf{50.9} & 28.5 & 40.2 & 40.8 & 14.7 & 44.9 & 63.8\\
DeNet-101 (wide) & 17 Hz & 33.8 & 53.4 & \textbf{36.1} & 12.3 & 36.1 & 50.8 & \textbf{29.6} & \textbf{42.6} & \textbf{43.5} & \textbf{19.3} & \textbf{46.9} & \textbf{64.3}
\end{tabular}}
\end{center}
\caption{MSCOCO average precision (AP) and average recall (AR) results evaluated on \texttt{test-dev2015} dataset.}
\label{table:mscoco}
\end{table*}


\subsection{Pascal VOC 2007}
We combined the \texttt{trainval} samples from Pascal VOC 2007 and 2012 \cite{pascal-voc} (denoted \texttt{07+12} in table) to produce 16,551 training samples. For testing we used Pascal VOC 2007 \texttt{test} containing 4,991 samples. We note that this dataset is considerably smaller than MSCOCO and therefore more susceptible to overtraining and image augmentation methods. Training time was \approxtilde 13 hours for DeNet-34 using $2\times$ Tesla P100s and \approxtilde 20 hours for DeNet-101 using $4\times$ Tesla P100s. In Table \ref{table:voc2007}, we provide the MAP and timing results. We observed the skip layer variant DeNet-34 improving upon SSD300's peak MAP by $1.6\%$ and 20Hz. In the near real-time domain DeNet-101 matches SSD512 at a higher evaluation rate. 




\begin{table}[tb]
\begin{center}
\begin{tabular}{ l|c|c|c}
 Model & Dataset & Eval. Rate & MAP \\
\hline
\rowcolor[gray]{.85} RPN (VGG) & 07+12 & 7 Hz & 73.2\% \\
\rowcolor[gray]{.85} RPN (ResNet) & 07+12 & 2 Hz & 76.4\% \\
\rowcolor[gray]{.85} R-FCN & 07+12 & 9 Hz & \textbf{80.5\%}  \\
Fast YOLO & 07+12& \textbf{155 Hz} & 52.7\%  \\
YOLO & 07+12& 45 Hz & 63.4\%  \\
SSD300 & 07+12& 58 Hz & 74.3\% \\
SSD512 & 07+12& 23 Hz & 76.8\%  \\
\hline
DeNet-34 & 07+12 & 83 Hz &  75.3\% \\
DeNet-34 (skip) & 07+12 & 82 Hz &  75.9\% \\
DeNet-101 & 07+12 & 34 Hz & 77.0\% \\
DeNet-101 (skip) & 07+12 & 33 Hz & 77.1\% \\
\end{tabular}
\end{center}
\caption{Pascal VOC 2007 mean average precision and timing. }
\label{table:voc2007}
\end{table}

\subsection{Pascal VOC 2012}
In this experiment we combine \texttt{trainvaltest} from Pascal VOC 2007 and \texttt{trainval} from Pascal VOC 2012 \cite{pascal-voc} (denoted \texttt{07++12} in table) to produce 21,503 training samples. Test scores are evaluated on 10,991 samples by the Pascal VOC 2012 testing server. 
For this dataset the DeNet-34 model matches SSD300, however, for reasons unknown, DeNet-101 demonstrates results below SSD512. For reference, we note that DeNet-101 obtains results near identical to the other ResNet-101 based model, RPN (ResNet) with an order of magnitude improvement in evaluation rate. Training time was \approxtilde 18 hours for DeNet-34 with $2\times$ Tesla P100s and \approxtilde 28 hours for DeNet-101  with $4\times$ Tesla P100s.

\begin{table}[tb]
\begin{center}
\begin{tabular}{l|c|c|c}
 Model & Dataset & Eval. Rate & MAP \\
\hline
\rowcolor[gray]{.85} RPN (VGG) & 07++12 & 7 Hz & 70.4\% \\
\rowcolor[gray]{.85} RPN (ResNet) & 07++12  & 2 Hz &  73.8\%\\
\rowcolor[gray]{.85} R-FCN & 07++12 & 9 Hz & \textbf{77.6\%} \\
YOLO & 07++12  & 45 Hz & 57.9\% \\
SSD300 & 07++12  & 58 Hz & 72.4\% \\
SSD512 & 07++12  & 23 Hz & 74.9\% \\
\hline
DeNet-34 & 07++12 &  \textbf{83 Hz} & 72.3\% \\
DeNet-101 & 07++12 & 33 Hz & 73.9\% 
\end{tabular}
\end{center}
\caption{Pascal VOC 2012 mean average precision and timing. }
\label{table:voc2012}
\end{table}

\section{Conclusion}
In this work, we describe a framework for sparse estimation with CNNs and present a novel region-of-interest detector and classification model which reduces manual engineering and improves state-of-the-art detection performance with real-time and near real-time evaluation rates. Utilizing deconvolution and skip layers first described in the context of semantic segmentation, we demonstrated a highly computationally efficient model with tightly coupled RoI, class prediction and bounding box regression. We provide further evidence that skip connections consistently improved detection rates for small and medium sized objects. While the \textit{wide} model variant highlighted the importance of corner map resolution for small and medium sized objects, and provides a natural pathway for future development. 

Analysis suggests our model performs particularly well when finer object localization is desirable. We propose that the improved localization is due to the much larger set of possible sampling bounding boxes that are feasible with our sparse sampling method i.e. $4.2\times 10^6$ compared to less than $2.5\times 10^4$ for SSD512 and RPN. This feature allows the model to potentially select a bounding box (before bounding box regression) which is significantly closer to the ground truth. Furthermore, since we no longer define a set of reference bounding boxes, this approach has reduced manual engineering requirements and can adapt well to problems which utilize bounding boxes with a very large range of aspect ratios and scales e.g. rotationally variant or non-rigid objects. 

\clearpage
\newpage

{\small
\bibliographystyle{ieee}
\bibliography{egbib}
}

\end{document}